\documentclass[a4paper]{article}
\usepackage[margin=25mm]{geometry}
\usepackage{amsmath}
\usepackage{amsfonts}
\usepackage{amssymb}
\usepackage[dvipsnames]{xcolor}
\usepackage{graphicx}
\usepackage{verbatim}
\usepackage{authblk}
\usepackage{hyperref}
\usepackage{graphicx}
\usepackage[authoryear]{natbib}
\immediate\write18{texcount -tex -sum  \jobname.tex > \jobname.wordcount.tex}

\providecommand{\keywords}[1]
{
  \small	
  \textbf{\textit{Keywords---}} #1
}

\title{Deep learning and differential equations for modeling changes in individual-level latent dynamics between observation periods}
\author[1,2]{G\"oran K\"ober} 
\affil[1]{Institute of Medical Biometry and Statistics, Faculty of Medicine and Medical Center – University of Freiburg, Germany}
\affil[2]{Freiburg Center for Data Analysis and Modelling, University of Freiburg, Germany}
\author[3,4]{Raffael Kalisch}
\affil[3]{Leibniz Institute for Resilience Research, Mainz, Germany}
\affil[4]{Neuroimaging Center, Focus Program Translational Neuroscience, Johannes Gutenberg University Medical Center, Mainz, Germany}
\author[3]{Lara Puhlmann}
\author[5]{Andrea Chmitorz}
\affil[5]{Faculty of Social Work, Education and Nursing Sciences, University Esslingen, Germany}
\author[4,6]{Anita Schick}
\affil[6]{Department of Public Mental Health, Central Institute of Mental Health, Medical Faculty Mannheim, Heidelberg University, Germany}
\author[1,2]{Harald Binder}

\begin{document}
\maketitle

\begin{abstract}
When modeling longitudinal biomedical data, often dimensionality reduction as well as dynamic modeling in the resulting latent representation is needed. This can be achieved by artificial neural networks for dimension reduction, and differential equations for dynamic modeling of individual-level trajectories. However, such approaches so far assume that parameters of individual-level dynamics are constant throughout the observation period. Motivated by an application from psychological resilience research, we propose an extension where different sets of differential equation parameters are allowed for observation sub-periods. Still, estimation for intra-individual sub-periods is coupled for being able to fit the model also with a relatively small dataset. We subsequently derive prediction targets from individual dynamic models of resilience in the application. These serve as interpretable resilience-related outcomes, to be predicted from characteristics of individuals, measured at baseline and a follow-up time point, and selecting a small set of important predictors. Our approach is seen to successfully identify individual-level parameters of dynamic models that allows us to stably select predictors, i.e., resilience factors. Furthermore, we can identify those characteristics of individuals that are the most promising for updates at follow-up, which might inform future study design. This underlines the usefulness of our proposed deep dynamic modeling approach with changes in parameters between observation sub-periods.
\end{abstract} \hspace{10pt}

\keywords{deep learning; dynamic modeling; longitudinal data; observational data; variable selection}

\section{Introduction}

Deep learning techniques, i.e., artificial neural networks with several layers, typically are associated with impressive performance on image data, such as in biomedicine \citep{esteva2017dermatologist}. However, there now also is a surge of proposals for combining deep learning with dynamic modeling, specifically using differential equations in a dimension-reduced latent representation obtained by neural networks \citep{rubanova2019latent, Chen2018NeuralODE, kidger2020neural}. In our own work with an application to psychological resilience research, we used such techniques for identifying individual-level temporal trajectories of mental health in relation to external stressor exposure \citep{koeber2020individualizing}. We could thus quantify resilience of individuals, understood as absent or moderate reactivity of mental health to stressor exposure when compared to individuals with similar levels of adversity \citep{kalisch2017resilience, 2021KalischKoeberFreshmo}, and use regularized regression techniques for identifying resilience factors, i.e., baseline characteristics that predict such resilient outcomes. Yet, this application also led to the need to allow for changes in resilience in observation sub-periods, i.e., changes in the individual-level differential equation parameters to thus accommodate the empirical observation that mental health reactivity to stressors does not necessarily constitute a stable trait but may change over time \citep{2021KalischKoeberFreshmo}.

While there are some other differential equation approaches for modeling dynamics and (normally-distributed) deviations from the main effects in psychological resilience \citep{montpetit2010resilience, driver2018hierarchical}, these do not allow for intra-personal changes of these dynamics. When not requiring modeling in a latent representation, i.e., when fitting dynamic models at the observed level, there are many potentially useful regression modeling frameworks \citep{rizopoulos2012joint, putter2017understanding}, and correspondingly various regression modeling approaches could be considered for updates of parameters. For example, L1-regularized regression could be considered 
\citep{2017InanPgeeRPackage, 2012WangPenalizedGEE, schelldofer2011estimation, schelldorfer2014glmmlassorpackage}
or coupling the likelihood of multiple points in time
\citep{schmidtmann2014coupled, zoller2016stagewise}, as proposed in our own work. Moreover, there are
many spline-based approaches that allow for modeling of changes in the dynamics
\citep{refisch2021datadriven, meng2021baysian, hong2012time, bringmann2017changing, wang2007groupSCAD}. To our knowledge, however,
there is no approach that allows for simultaneously identifying a mapping to a latent
space for dimension reduction and changes in individual-level dynamics with differential equations.

In contrast to such regression modeling frameworks, we consider modeling with ordinary differential equations (ODEs) in a dimension-reduced latent representation obtained by artificial neural networks, specifically variational autoencoders (VAEs; \citep{kingma2013auto}). We propose an approach for estimating separate sets of differential equation parameters for the two intra-individual sub-periods in our application. We acknowledge potential intra-individual similarity between the sub-periods by tying the parameters together with a penalty term, where estimation is enabled by differentiable programming \citep{Innes2019DiffProgramming, hackenberg2021using}. We subsequently use the L1-regularized regression, i.e., the Lasso, to identify predictors of resilience (that is, in approximation, stressor reactivity) in the in the sub-periods and thus characteristics of individuals where follow-up measurement might be valuable. 

In the following, we briefly describe the psychological resilience application that motivates our methods development in Section \ref{marp-data} before describing the proposed approach in Section \ref{methods} and illustrating it with results from the application in Section \ref{results}. We subsequently discuss the potential usefulness of our approach in other application settings and potential further extensions. 

\hypertarget{marp-data}{%
\section{A psychological resilience application of latent dynamic modeling}\label{marp-data}}

Psychological resilience is the maintenance or rapid recovery of a healthy mental state during and after times of adversity \citep{kalisch2017resilience}. One element of the definition is that both mental health problems (P) and stressor exposure (E) can change over time and may even do so permanently. Influential resilience studies \citep{bonanno2011resilience} investigated how mental health changes in response to one single potentially-traumatic life event and found groups of similar individual mental health trajectories such as resilient (showing stably good or improving mental health in the months or years after the event) or vulnerable (showing stably poor or worsening mental health). These studies implicitly assume that the observed temporal changes in mental health are due to only a single stressor event and that individual differences in the mental health trajectory can be explained by some baseline individual characteristic. However, most individuals are continuously exposed to more or less severe stressors. These may include macrostressors (severe life events) but also more ``mundane'' microstressors, or daily hassles \citep{hahn1999daily}, which also have an impact on mental health \citep{serido2004chronic, 2021KalischKoeberFreshmo}. Further, when exposed to hardships, individuals may undergo processes of learning and adaptation that may make them more or less reactive to such mental health challenges. This is supposed to express in changes in the strength or efficiency of the predictive resilience factors, in turn predicting changes in mental health reactivity. For example, an increase in someone's emotion regulation capacity (a resilience factor), as can sometimes be observed in individuals undergoing difficult life phases, may translate into reduced reactivity and eventually better mental health outcomes (a resilience process).  In the maladaptive case, a breakdown in emotion regulation may lead to worse outcomes \citep{kalisch2019deconstructing, 2021KalischKoeberFreshmo}. From this perspective, resilience research is a paradigmatic example of applications that require a dynamic analysis of both predictors and outcomes. In short, investigating resilience should ideally involve repeated longitudinal measurements of stressors and mental health. Further, as far as predictive individual characteristics may also change, resilience studies should ideally also assess such potential resilience factors repeatedly \citep{kalisch2017resilience, kalisch2019deconstructing}. Such longitudinal studies pose manifold problems for data collectors and analysts. 

The Mainz Resilience Project (MARP) is an ongoing study that started in
2016 with a planned study duration per participant of seven years. MARP
is conducted by the University Medical Center Mainz and the Leibniz
Institute for Resilience Research \citep{kampa2018combined}. Our choice of two sub-periods is motivated by the design of the MARP study, where potential resilience factors (predictors) are repeatedly measured approximately every 1.75 years in laboratory battery time points (B0, B1, ...), whereas repeated online measures of stressor exposure and mental health problems, serving to determine individual stressor reactivity (outcome), are regularly conducted at higher frequency (every three months, time points T0, T1, ...) at and between the battery time points \citep{2021KalischKoeberFreshmo}. At the current stage of ongoing data collection, B0 can be used to predict stressor reactivity between B0 and B1 (sub-period 1: T0--T6) and B1 to predict reactivity afterwards (sub-period 2: T6 and later). 

For inclusion into longitudinal modeling there must be at least two observations, i.e., online measurements at different time points T, of both P and E for an individual. This is the case for \(N=N_{sp1}=181\) respondents in a first sub-period of 1.75 years (terminated by the second resilience factor battery B1). In subsequent sub-period after B1 (corresponding to the online measurements T6 and later), data of \(N_{sp2}=120\) respondents are available. On average, \(11.2\) observations (\(min = 2\), \(max = 23\), \(sd = 6.1\)) were recorded. This corresponds to a total of \(N_{E} = 2.052\) and \(N_{P} = 2.087\) observations in the two sub-periods. Each observation at T comprises \(p=28\) mental health items (GHQ-28; \citep{goldberg1997validity}) and \(e=58\) daily hassle items (MIMIS battery; \citep{chmitorz2020microstressors}). Each mental health item is scored by the participants on a severity scale ranging from 0 to 3; for each daily hassle item, participants indicated the number of days in the last week (0 to 7) at which the hassle occurred. Participants were recruited in a critical life phase aged 18 to 20 at study inclusion with a prehistory of critical life events.

\begin{figure}
\centering
\includegraphics[width=\textwidth]{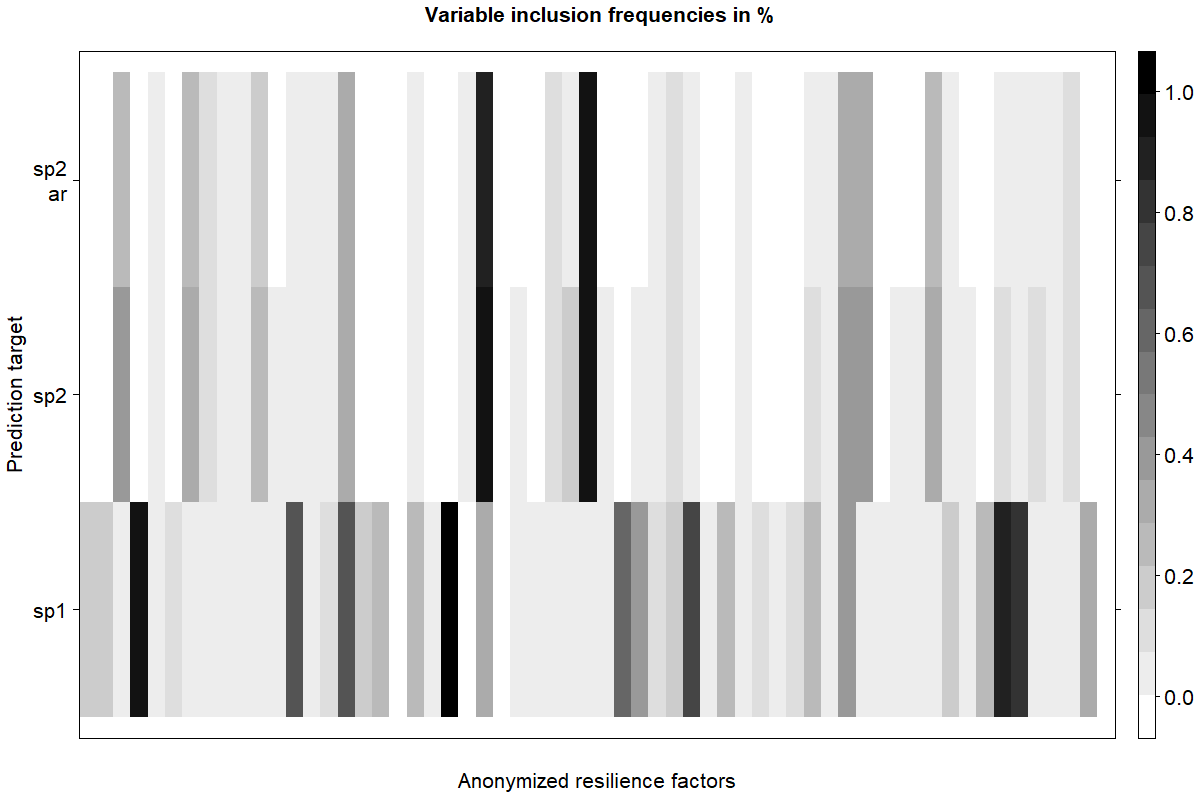}
\caption{Variable inclusion frequencies (VIF) of the Lasso analysis with
different prediction targets. This heat map provides insights into which
battery measures (x-axis) predict the targets (y-axis) derived for this
longitudinal focus. All rows of the y-axis depict the two
sub-periods or, more precisely, \(\eta_{i,2,sp1}\) and \(\eta_{i,2,sp2}\) from the ODEs (see Equation \eqref{eq:ODEs}),
which captures the individual gradient of mental health in response to
stressor exposure in the dynamical system. Importantly, sp1 was predicted solely with B0 (initial lab visit at the beginning of the study). The second sub-period sp2 was predicted with both B0 and B1 (the
latter is the first repetition of the initial lab visit's testing
battery around 1.75 years later), albeit with unequal penalty weights.
This plot indicates that there are five stable and predictive variables
in B0 (see lowest row) when predicting sp1. Due to their high VIF, these variables are penalized with
zero/very small \(w_{sp2}\) when predicting sp2 (see the dedicated
\protect\hyperlink{a-two-stage-lasso-approach-to-repeated-baseline-updates}{Methods}
section for more details). Accordingly, they are almost certainly
entailed in the prediction of sp2 dynamics as well. Therefore, in the two upper rows, we show only the B1 predictors. Regarding potential resilience factors in the second testing battery B1, there is a mixed
picture. While updates of already picked variables are chosen rather
seldom, other measures now play a more important role when predicting sp2. The upper
row (sp2 ar, for autoregressive) includes \(\eta_{i,2,sp1}\) as an additional predictor in the Lasso analysis. This mitigates the VIF only slightly, indicating a rather stable pattern of particularly valuable parameter updates in B1. We deliberately do not show or discuss the specific
constructs that we found to predict the individual dynamics since we
focus on methods development here.\label{vif}}
\end{figure}

In addition to longitudinal measurements (T0, T1, ...), there is an extensive battery of potential resilience factors for characterizing individuals. These repeated resilience factor testing batteries (B0, B1, ...) comprises questionnaire-based, neuropsychological, neuroimaging, and biological assessments, of which we here focus on the questionnaires. This information, which will be used for prediction modeling with the Lasso, is available for (\(N_{lasso, sp1} = 120\)) at B0, and (\(N_{lasso, sp2} = 53\)) respondents with renewed assessment at B1. This particular data situation allows us to investigate interesting downstream tasks, e.g., which of the repeatedly assessed potential resilience factors at B1 are worth updating the most to predict (changes in) the dynamics of stressor reactivity (see Figure \ref{vif}).

\hypertarget{methods}{%
\section{Methods}\label{methods}}

\begin{figure}
\centering
\includegraphics[width=\textwidth]{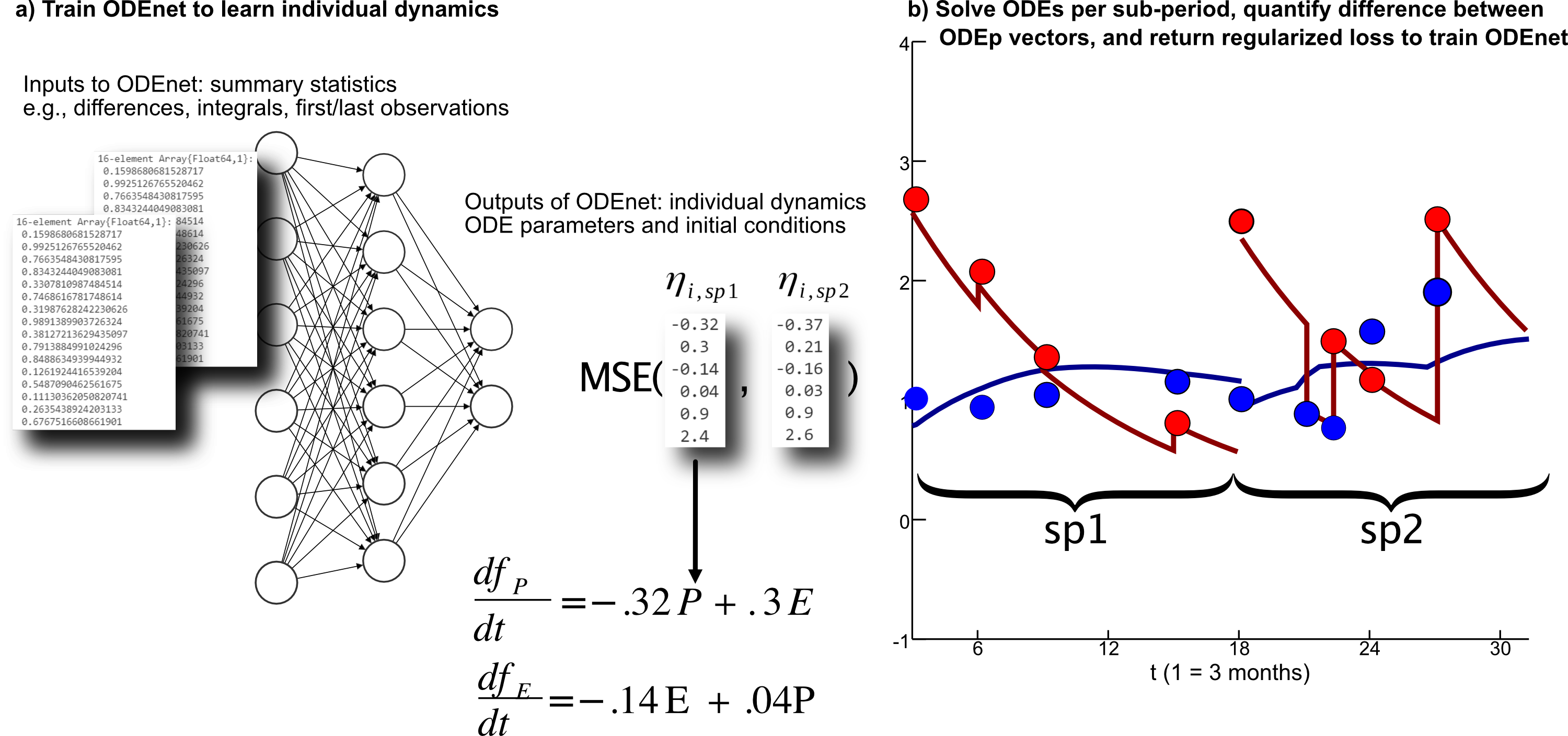}
\caption{Overview of ODEnet extension for intra-individual learning of potentially changing dynamics. We parameterize our system of ordinary differential equations (ODEs) with a feed-forward neural network (ODEnet) which receives summary statistics of the respective sub-period (sp1 or sp2) as inputs. The critical advancement of our algorithm is the distinction and intra-individual coupling of sp1 and sp2 for each respondent. The mean squared error (MSE) of each intra-individual parameter set is summed, weighted by \(\lambda_{sp}\), and provided as additional term to the loss function (see Equation \eqref{eq:loss_odenet}). Sp1 covers all observations T0--T6 from the baseline battery measurement (B0) to the repeated resilience factor testing battery (B1). Sp2 comprises all observations of mental health and stressor exposure after that. \label{diagram}}
\end{figure}

We first provide a brief overview of the proposed approach, before subsequently describing its components in detail. A schematic illustration is given in Figure \ref{diagram}. VAEs are used for separate dimensionality reduction of mental health and stressor items. Longitudinal summary statistics of the time series of the resulting latent representation are fed into a second neural network---the ODEnet---which provides individual-level ODE parameters and their initial conditions (ICs) as output. The ODE parameters (\(\eta_{i,1,sp}, \ldots, \eta_{i,4,sp}\)) and the ICs are obtained for each individual, separately for the two sub-periods. We introduce individual-level dependency by
penalizing the difference of each ODE parameter between the sub-periods
(but not the ICs). Thereby, the ODEnet is trained to find
ODE parameters that best describe the dynamics of the respective
sub-period and, at the same time, enforce some correlation between
parameter sets from the same respondent, reflecting that the data is
provided from the same respondent.

\hypertarget{dimensionality-reduction-per-time-point-with-vaes-for-count-data}{%
\subsection{Dimensionality reduction per time point with VAEs}\label{dimensionality-reduction-per-time-point-with-vaes-for-count-data}}

We choose Variational Autoencoders \citep[VAEs; ][]{kingma2019intro} to find lower-dimensional latent distributions \(q_{\phi}(z|x)\) making use of variational inference \citep{blei2017variational}. Let \(x^P_i = (x^P_{i,1}, \ldots, x^P_{i,p})\), be a vector of p items of P and \(x^E_i = (x^E_{i,1}, \ldots, x^E_{i,e})\) be a vector of e items of E and \(i \in \{1,\ldots, N_P\}\) or \(i \in \{1,\ldots, N_E\}\), respectively (see section \ref{marp-data}). The latent variables of E or P are denoted as \(z^{E}_{i}\) and \(z^{P}_{i}\) for each realized observation \(i\). Then, the VAEs are trained by maximizing the evidence lower bound (ELBO) of the marginal log-likelihood

\[
  \begin{split}
    \log p(x_1,\ldots, x_N) & = \sum_{i=1}^{N} \log p(x_i) \\
              & \ge \sum_{i=1}^{N} (\mathbb{E}_{q(z|x)}  \log p(x_i|z)  - \text{KL }(q(z|x_i)||p(z)))
  \end{split}
\]

where the first term of the right hand side is the expectation of the
log-likelihood of \(x_i\) given \(z\) with respect to \(q(z|x)\). The
Kullback-Leibler divergence (\(D_{KL}\)) penalizes deviations of the
posterior from the prior. Temporal dependence between the observations
is established by the system of ODEs. 

Intuitively, VAEs comprise an encoder and a decoder. The purpose of training the encoder (aka inference or recognition model) is to find good
variational parameters \(\phi\) that parameterize the distributions \(q_{\phi}(z|x)\) in the latent space in a way, the decoder can reconstruct samples of that latent distribution that resemble the inputs \(x_i\). In our case, the former is achieved by two separate \(\text{EncoderNets}\)

\begin{align}   
  (\mu^P_i, \log\sigma^P_i) &= \text{EncoderNet}_{\phi_P}(x^P_i) \label{eq:encodernetP}\\
	(\mu^E_i, \log\sigma^E_i) &= \text{EncoderNet}_{\phi_E}(x^E_i) \label{eq:encodernetE} 
\end{align}

which yield the latent distributions 

\begin{align} \label{eq:encoderdist}
  q_{\phi}(z^P_i|x^P_i) &= \mathcal{N}(\mu^P_i, \log\sigma^P_i) \\
	q_{\phi}(z^E_i|x^E_i) &= \mathcal{N}(\mu^E_i, \log\sigma^E_i)
\end{align}

for each complete observation \(i\). The two separate \(\text{DecoderNets}(\theta)\) then take samples of \(q_{\phi}(z|x)\) and 
are trained to increase the log-likelihood of the inputs given samples of the posterior distributions. 
Please note, while the expected value of the posterior distribution \(\mu\) captures the position of the
observation in the latent space, the variance \(\sigma\) captures the uncertainty
that is related to this mapping \citep[see also][]{kingma2019intro}.

For computation, we plug in the
Poisson log-likelihood and the closed form of \(D_{KL}\) for a Gaussian
prior and posterior. Thereby, training objective of our VAEs become

\[
  \begin{split}
    \text{Loss}(\theta, \phi; x_i) = & \underbrace{\frac{1}{2} \sum_{i=1}^N (1+\log((\sigma(x_i))^2) - (\mu(x_i))^2 - (\sigma(x_i))^2)}_{D_{KL}(\phi)} + \\
    & \underbrace{\sum_{i=1}^N (\lambda_i - x_i \times \log(\lambda_i))}_{\text{Poisson reconstruction error }(\theta)} + \\
    & \underbrace{\lambda_{VAE} \times (\sum_{i}\phi + \sum_{i}\theta)}_{\text{Weight regularization}}
  \end{split}
\]

where \(\mu\) and \(\sigma\) are the mean and standard deviation of the
latent distribution depending on the observed values \(x_i\).
The expected value of the Poisson distribution of each item is denoted as \(\lambda_i\) depending on the observation. 
To prevent overfitting, we regularized the encoder
(\(\phi\)) and decoder (\(\theta\)) weights of both VAEs with
\(\lambda_{VAE}\).

The Poisson distribution does not perfectly match the
\protect\hyperlink{marp-data}{MARP data} since we find moderate levels
of overdispersion for some items. Yet, the Poisson distribution avoids
an additional parameter, which might be difficult to determine, by
coupling the expected value and the dispersion. We used the \(tanh\)
activation functions in the middle layers. In the final layer, we used a
\(ReLU\) activation function to strictly pass non-negative values to
\(\lambda_i\).

\hypertarget{ode-system-to-model-resilience-trajectories-in-the-latent-space}{%
\subsection{ODEs to model the trajectories of E and P in the latent
space}\label{ode-system-to-model-resilience-trajectories-in-the-latent-space}}

We use ODEs to couple the latent representations of P and E over time and allow for separate intra-individual parameter sets for periods between the resilience factor testing batteries (B0, B1), providing potential predictors of the individual dynamics. 
The exact design of such an ODE system is a crucial modeling decision since it governs how each component changes and requires domain expertise. We were able to model a slightly more
complicated system of ODEs than \citet{koeber2020individualizing}.
Specifically, we use

\begin{align}\label{eq:ODEs}
  \frac{df_{z^{P}_{i,t}}}{dt} &= \eta_{i,1,sp} \times z^{P}_{i, t} + \eta_{i,2,sp} \times z^{E}_{i, t} \\
  \frac{df_{z^{E}_{i,t}}}{dt} &= \eta_{i,3,sp} \times  z^{E}_{i, t}  + \eta_{i,4,sp} \times z^{P}_{i, t}
\end{align}

where changes in \(z^{P}_{i, t}\), the latent trajectory value of mental
health problems (P) of respondent \(i\) at \(t\) with \(i \in \{1, \ldots, N\}\) and \(t \in \{1, \ldots, T\}\), 
and \(z^{E}_{i, t}\), the latent trajectory value of 
stressors exposure (E), are driven by their own current value.
Additionally, \(z^{P}_{i, t}\) is allowed to change in response to \(z^{E}_{i, t}\) and vice versa.
Subscript \(sp\) in \(\eta_{i,1,sp}\) indicates that each individual set of parameters 
differs intra-individually between sp1 and sp2.

Negative values for \(\eta_{i,1,sp}\) and \(\eta_{i,3,sp}\) effectively
realize system-inherent damping \citep{boker2010modeling}, where high E
and P values are more quickly driven back to low values in the latent
space. Thus, a high negative \(\eta_{i,1,sp}\), in particular, reflects
good recovery from mental health problems (since the majority of
observations are mapped into the positive valued latent space or close
to zero). This can be understood as one facet of stressor reactivity, where possible surges in mental health problems tend to be short-lasting. Positive values for \(\eta_{i,2,sp}\) realize the adverse
effect of E on P. A low positive \(\eta_{i,2,sp}\) value thus reflects low responsivity of mental health to stressors in the first place, as the second element in our equation system that, overall, describes an individual's stressor reactivity. \(\eta_{i,4,sp}\) allows for the opposite direction and expresses that
people with high P report more E in the future. This is deemed plausible, both because mental health problems are stressors in their own right that can induce further adverse reactions and because mentally ill persons may generate, or be confronted with, more conflicts or other types of adversities \citep{gerin2019heightened}. Key parameters, however, for the interpretation of an individual's resilience status are \(\eta_{i,1,sp}\) (recovery) and \(\eta_{i,2,sp}\) (reactivity).

At each realized measurement of stressor exposure, its integrator
\(f_{z^{E}_{i}}\) is updated to the mapping of the actually observed
values to the latent space \(z^{E}_{i}\) at the precise point in (study) time 
when this observation was taken. This reflects the notion
that stressor exposure levels are only partly driven by an endogenous property of
the ODE system (i.e., damping and mental health) but mainly reflect
exogenous forces, that is, the sudden occurrence or absence of stressors
that lead to abrupt changes in the latent stressor values (see above).

The benefits of such an ODE system compared to discrete-time models like
regression are crucial for analyzing the data from the MARP study. Most
importantly, differential equations take all available information at
the precise time into account. Thereby, irregular sampling intervals and
entirely (or partly) missing observations are dealt with by the
properties of our dynamical system (assuming non-informative
missingness).

\hypertarget{individual-and-potentially-changing-ode-parameters}{%
\subsection{Finding individual and potentially changing ODE
parameters with the ODEnet}\label{individual-and-potentially-changing-ode-parameters}}

The critical advancement of our method is the extension of the ODEnet to learn potentially 
changing ODE parameters \(\eta_{i,s,sp}\). The key idea is depicted in Figure \ref{diagram}. 
The ODEnet is trained by minimizing the \(\text{Loss}_{ODE}(\tau, \eta_{i,s,sp})\) and provides a two-element 
vector of ODE parameters, each of length \(s \in \{1, \ldots, 4\}\), for both sub-periods

\[
	\eta_{i,s,sp} = (\eta_{i,s,sp1}, \eta_{i,s,sp2}) = \text{ODEnet}_\tau(x^P_i, x^E_i)
\]

with the observed items \(x^P_i\) and \(x^E_i\) as inputs and the trainable parameters \(\tau\).
The ODEnet internally calls the EncoderNet (see Equation \ref{eq:encodernetP} and \ref{eq:encodernetE}) which maps the observed values into the latent space. 
Subsequently, we calculate several summary statistics of the intra-individual sub-period in the latent space (e.g., integrals, first and last observations, 
and differences; see Table 1 in \citet{koeber2020individualizing}).
It was left to gradient descent to find a good combination of these summary statistics to minimize \(\text{Loss}_{ODE}(\tau, \eta_{i,s,sp})\).
All summary statistics are required to be computable with only two observations from P and E (which can be reported at any point in time, not necessarily during the same observation). 
The main purpose of the ODEnet is to minimize the sum of
the squared difference of the trajectory \(f_{z_i}(t, \eta_{i,s,sp})\) and the
mean of the latent space distribution \(\mu_i\) at the precise
point in study time \(t\) depending on period-specific individual ODE parameters \(\eta_{i,s,sp}\). 
Accordingly, the training objective of the ODEnet is

\begin{equation} \label{eq:loss_odenet}
\begin{split}
  \text{Loss}_{ODE}(\tau, \eta_{i,s,sp}) = & \sum_{i=1}^{N}\sum_{t=1}^{T} (f_{z_i}(t, \eta_{i,s,sp}) - \mu_{z_{i,t}})^2 \quad + \\
  & \lambda_{sp} \times \sum_{s=1}^4(\eta_{i,s,sp1}-\eta_{i,s,sp2})^2 \quad  + \\
  & \lambda_{ODEp} \times \sum_{s=1}^4\eta_{i,s,sp}\ \quad + \\
  & \lambda_{ODEnet} \times \sum_{i}\tau \text{.}
\end{split}
\end{equation}

Importantly, the period-specific individual parameter sets within \(\eta_{i,s,sp}\)
are tied together by penalizing the squared difference of the two parameter sets \(\eta_{i,s,sp1}\) and \(\eta_{i,s,sp2}\). 
This bond of ODE parameters across the sub-periods reflects their intra-individual dependence. Furthermore, this tying helps to
prevent overfitting of the ODE parameters to a certain 
period with, e.g., only a few unusual measurements. The hyperparameter \(\lambda_{sp}\)
\(\in \mathbb R_{> 0}\) allows to tune the strength of this connection. We
additionally penalize both the sum of the ODE parameters with
\(\lambda_{ODEp}\) as well as the weights of the ODEnet with
\(\lambda_{ODEnet}\) to avoid overfitting.

The concrete values of the hyperparameter \(\lambda_{sp}\) can be decided based on subject-matter
considerations, e.g., by imposing stronger similarity between
intra-individual sub-period dynamics for subject areas which are known
for evolving slowly, or by optimization criteria, e.g., cross-validated
mean squared error (MSE) as usual in Lasso analyses \citep{hastie2015statistical}. Please
note, the differences of the initial conditions (IC) are not penalized,
although the ODEnet also provides them.

More detailed, the ODEnet is a three-layer feed-forward neural network. We use a \(ReLU\) activation function in the middle layer and no
transformation in the final layer. Using no activation function in the last layer does allow the ODEnet to find the parameters of the dynamical system freely; 
put differently, it enables the ODEnet to provide the full numerical range of possible ODE parameters. 
We increased the capacity of the ODEnet---the mid-layer has 12 nodes now---compared to
\citet{koeber2020individualizing}, mainly due to the increased number of
training examples (with the additional split at the second lab visit,
most respondents contributed two time series). Hyperparameter tuning on
the empirical data with regularization and a simulation study agreed
that this architecture balances bias and variance well. Learning the ICs
overcomes the limitation of \citet{koeber2020individualizing}, which
regarded the first observation as ground truth.

To increase training stability, we initialized the ODEnet with very small weights and trained it rather slowly for 100 epochs with a small learning rate of \(\alpha = 10e^{-4}\). 
Furthermore, we scale the inputs to ensure approximately equal
numerical size. Flexible dynamical modeling is provided by 
{\tt{DifferentialEquations.jl}} \citep{Rackauckas2019DifferentialEquations} and
differentiable through neural nets via {\tt{DiffEqFlux.jl}}
\citep{Rackauckas2019DifferentialEquations}. To deal with unit non-response,
\(\text{Loss}(\theta, \phi; x, \lambda_{VAE})\) and
\(\text{Loss}_{ODE}(\tau, \eta_i)\) are only evaluated at actual
measurement time points. All neural networks were trained with {\tt{Flux.jl}}
\citep{FluxJl} and the {\tt{Adam}} optimizer \citep{kingma2014adam}. We used
this \href{https://alexlenail.me/NN-SVG/index.html}{website}
\citep{lenail2019nn} as a basis for drawing the ODEnet in Figure
\ref{diagram}.

\hypertarget{a-two-stage-lasso-approach-to-repeated-battery-updates}{%
\subsection{A two-stage Lasso approach to repeated battery updates}\label{a-two-stage-lasso-approach-to-repeated-battery-updates}}

We assess the capabilities of the incoming predictor information to
improve the prediction beyond the older (but potentially still relevant)
subject information with a variant of the Lasso tailored to this data
situation. The particularity of this data is that an extensive (and expensive) multi-modal resilience factor testing battery including---next to questionnaires---also neuroimaging and biosampling, amongst others, is repeated in regular intervals (at B0, B1, ...). In the meantime, several samples of key
resilience outcome variables (mental health problems and stressor exposure) are
drawn (at T0, T1, ...). To be able to learn about the usefulness of the incoming testing
battery information (B1) for prediction, we add an additional weight vector
\(w\) to the standard Lasso training criterion. The optimal parameters
for all available battery information (B0 and B1) predicting the dynamics in the second sub-period (sp2) are 

\[
 \hat{\beta}_{sp2} =  \underset{\beta}{\operatorname{argmax}} \{ \frac{1}{2} |y - X \times \beta|^2 + \lambda_{\text{Lasso}} \sum^p_{j=1} w \times \beta \} 
\]

where \(X\) is the design matrix, y is a resilience-related outcome, and \(\lambda_{\text{Lasso}}\) is the penalty coefficient of the Lasso.

This approach has been already suggested for cross-sectional data
settings before \citep{zou2006adaptive} and can be implemented simply
using the established R package {\tt{glmnet}} \citep{friedman2010glmnet}; this
circumstance fosters reusability of the approach, also independently of
the estimation of dynamics with the ODEnet as suggested above. In the
first step of our two-stage approach, the Lasso is trained only with
predictor data of the initial baseline observation (B0) to predict the
learned dynamics of the first sub-period. Since no previous knowledge is
assumed, the penalty factor is set to its default (\(w = 1\)) for all
potential predictors. In a second step, both battery information from B0 and B1
provide potential predictors of the dynamics of
the second sub-period (sp2). We use the previous Lasso analysis and penalize
the initial data from B0 with \(-\log(VIF)\) while we penalize incoming
data uniformly with the default (\(w=1\)).

We use the variable inclusion frequencies (\(VIF\)) because single lasso
analyses might be unstable regarding the selected variables. For this
reason, \citet{wallisch2021stability} suggest using resampling to
determine model stability (see also \citet{heinze2018variable} and
\citet{sauerbrei2015stability}). More precisely, we resampled a
proportion of \(m=0.8\) and repeated the Lasso analysis 1.000 times,
using 6-fold cross-validation to determine the strength of the lasso
penalty once.

More detailed, assuming that each testing battery (B0, B1, ...) provides the same
number of variables \(v\), and that there are less individuals who met the
minimum requirements, \(N_{sp1}\) is usually larger than \(N_{sp2}\). The \(v \times n_{sp1}\) matrix
\(X_{sp1}\) of potential resilience factors is accompanied by
the weight vector \(w_{sp1} = (w_{sp1, 1}, \ldots , w_{sp1, v}) = (1, \ldots, 1)\) when
predicting an indicator of the dynamics of the first sub-period
\(y_{sp1}\). For the second sub-period, however, we stack both baseline
information on top of each other, accordingly, \(X_{sp2}\) is
\(m \times n_{sp2}\) where \(m = 2 \times v\) (in case
\(v_{sp1} = v_{sp2}\)). The weight vector of the second sub-period can be
expressed as

\[
    w_{sp2} = \left\{\begin{array}{lr}
        -\log(VIF(X_r)), & \text{for } r \leq \frac{m}{2}\\
        1, & \text{for } r > \frac{m}{2}\\
        \end{array}\right\}
\]

with \(r \in \{1, \ldots, p\}\). 

We predict the ODE mental health responsivity parameter \(\eta_{i,2,sp}\) which is a choice made for this paper, intending to demonstrate the method. One could also choose a different prediction target, e.g., \(\eta_{i,1,sp}\) which reflects the recovery of mental health, or impose a considerable amount of stress on the entire individual dynamical system (or, more precisely, each learned parameter set \(\eta_{i,s,sp}\) plugged into the shared dynamical system structure) and predict mental health at predefined time points (as done by \citet{koeber2020individualizing}), depending on which aspect one thinks is more interesting from a subject-matter perspective.

\hypertarget{results}{%
\section{Results}\label{results}}

\begin{figure}
\centering
\includegraphics[width=\textwidth]{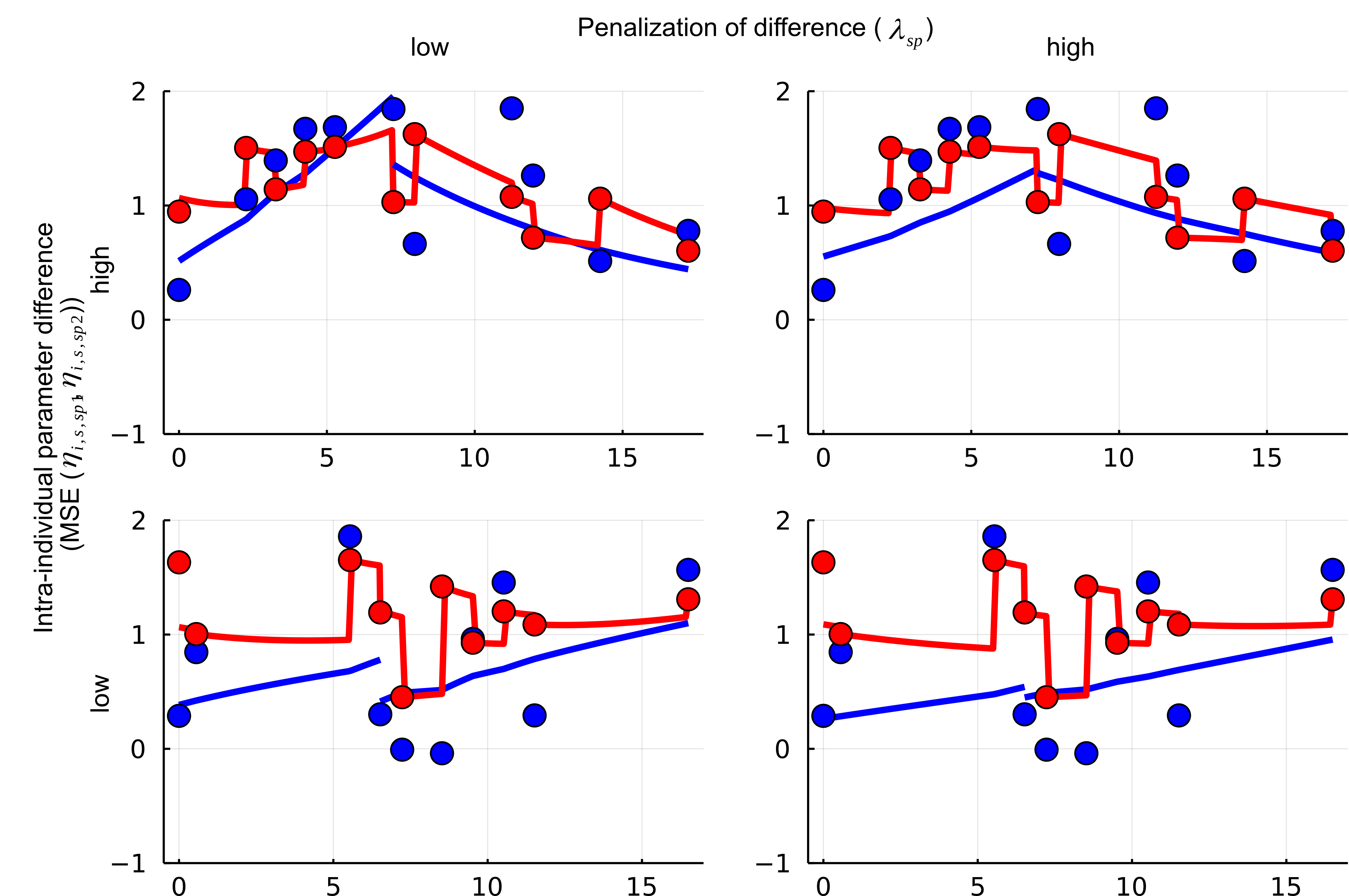}
\caption{This figure shows an exemplary comparison of the differences between sub-periods of dynamical systems. Mental health problems (P, blue) and stressor exposure (E, red) are shown for two respondents (rows) with low (\(\lambda_{sp} = 0.4\)) and high (\(\lambda_{sp} = 3.6\)) values of intra-individual deviation penalty (columns).  The upper row shows a respondent with considerable change in the ODE parameters between sub-period 1 to 2 (sp1 and sp2). P increase in sp1. In contrast, P decreases in sp2. This change is visibly mitigated, although still present, when the ODEnet is trained with a stronger penalization of parameter differences \(\lambda_{sp} = 3.6\) (right column). The second row shows a respondent with a low change in the ODE parameters. Accordingly, increasing \(\lambda_{sp}\) (from left to right) does not affect the intra-individual comparison strongly. \label{comp}}
\end{figure}

First, we compare dynamical systems (see Equation \eqref{eq:ODEs}) of mental health problems (P, blue) and stressor exposure (E, red) of two exemplary individuals (rows) with different intra-individual difference penalization terms \(\lambda_{sp}\) (columns; see Equation \eqref{eq:loss_odenet}) in Figure \ref{comp}. As in Figure \ref{diagram}, the y-axis shows the position of P and E in the latent space, and the x-axis shows time (where one unit in x represents three months in study time).  The expected values of the latent value distributions, learned by the VAEs, are expressed as dots. The trajectories, governed by the ODEs, are depicted as lines. Figure \ref{comp} exemplifies that this algorithm is able to detect potential sub-period differences and that \(\lambda_{sp}\) can alter the strength of this difference (based on, e.g., subject-matter considerations or cross-validation).

Furthermore, this particular data situation allows us to investigate the practice-oriented question of which parts of the repeated battery are particularly worth updating. To illustrate this, we predict the ODE mental health responsivity parameter \(\eta_{i,2,sp}\)
directly---which captures how the level of stressor exposure influences
the gradient of mental health problems---with a two-staged lasso
approach (see the 
\protect\hyperlink{a-two-stage-lasso-approach-to-repeated-baseline-updates}{Methods}
section for more details) and show and discuss the results in Figure
\ref{vif} with a heat map. 

Please note that while only information of B0 is included in the
analysis of sp1, variables of B0 and B1 are included in the analysis of
sp2. However, due to the very low penalization weight \(w\) of the
frequently selected variables (and reversely the very high weight of the
non-selected variables) they are almost certainly (not) selected, which
renders their visualization pointless. Since variable selection approaches
are known for being unstable, we choose VIF \citep{wallisch2021stability} for both as the decisive criterion for our two-staged procedure determining the weights and for
visualizing the results.

A subsequent question is whether we can predict the changes of \(\eta_{i,2,sp2}\) in an auto-regressive manner. Put differently, do the recommendations on promising parameter updates hold even when including \(\eta_{i,2,sp1}\) into the Lasso? All findings are discussed in the figure captions.

\hypertarget{discussion}{%
\section{Discussion}\label{discussion}}

Recently, there have been several proposals for deep dynamic modeling, where differential equations are used on latent representations obtained by VAEs. Motivated by an application from psychological resilience research, we removed the
assumption of constant dynamics throughout the whole study
period, implicit in these approaches. Instead, we showed how individual-level dynamic models could be fitted for two individual sub-periods and introduced a dependency between by a penalization term, which made our approach feasible also for a dataset with relatively small number of individuals.

We then used regularized regression, specifically the Lasso approach, to identify potential resilience factors from a baseline battery and its follow-up update that could predict summary statistics obtained from individual resilience trajectories. Such a variable
selection approach, e.g.,  allows pruning the battery for subsequent lab visits, thus potentially saving time and money. We find that five resilience factors in the initial testing battery are
particularly promising for predicting the dynamics in the first sub-period. While
we pick few variables from the second lab visit in the second sub-period  again, we could identify a small set variables that could be recommended for update. This suggests that a dramatically reduced battery would be sufficient
for this prediction task. This findings holds when explicitly predicting parameter changes with an auto-regressive approach.

To deal with a relatively small number of respondents, our model also made efficient use of the vast majority of training data to learn the
lower-dimensional representations, trajectories, and predictors. In particular, we provided all measurement time points separately as observations to the VAEs, and did not attempt to fit more complex neural networks architectures with time structure, such as recurrent neural networks, as time structure is already covered by the ODEs. Such dynamic models at the heart of neural networks are known to reduce sample size requirements in deep learning algorithms drastically \citep{rackauckas2020universal}. 

We also had to take several additional countermeasures to reduce the data hunger of deep learning methods, such as coupling the expected value and dispersion parameter by assuming a Poisson distribution for the item data. Furthermore, we have trained our neural networks on all available data and did not reserve any test set to investigate the generalization error. Instead, the role of subsequent validation is provided by the prediction with the Lasso. Specifically, the argument is that we could not have identified characteristics that prediction resilience prediction targets of the dynamic modeling would not reflect true underlying structure. Our insights and approaches regarding the feasibility of deep learning 
approaches when facing moderate sample sizes may also be more generally helpful in other studies that want to use similarly sophisticated approaches but are in doubt of whether
their sample size is sufficient. 

\renewcommand{\abstractname}{Acknowledgements}
\begin{abstract}
 This project has received funding from the European Union’s Horizon 2020 research 
    and innovation programme under grant agreement No. 777084 (DynaMORE project).
\end{abstract}
\vspace*{1pc}

\renewcommand{\abstractname}{Conflict of Interest}
\begin{abstract}
    RK receives advisory honoraria from JoyVentures, Herzlia, Israel.
    The authors declare that the research was conducted in the absence of any
    commercial or financial relationships that could be construed as a potential
    conflict of interest.
\end{abstract}

\bibliography{my}
\end{document}